\documentclass[letterpaper, 10pt, conference]{ieeeconf}
\IEEEoverridecommandlockouts
\usepackage{graphicx} 
\usepackage{float}
\usepackage{amsmath}
\usepackage{threeparttable}
\usepackage{multirow}
\usepackage[misc,geometry]{ifsym} 

\usepackage{epstopdf} 
\usepackage{epsfig}
\usepackage{mathptmx} 
\usepackage{times} 
\usepackage{amsmath} 
\usepackage{amssymb}  
\usepackage{mathtools}
\usepackage{cite}
\usepackage[final]{pdfpages}
\usepackage{color} 
\usepackage{siunitx} 
\usepackage{dblfloatfix}
\let\DeclareUSUnit\DeclareSIUnit

\DeclareUSUnit\inch{"}

\usepackage{multicol}
\usepackage{multirow}
\usepackage{array}

\usepackage[utf8]{inputenc}
\usepackage{comment} 
\usepackage{array}

\usepackage{hyperref}

\title{\LARGE \bf Computational Design and Fabrication of Corrugated Mechanisms from Behavioral Specifications}

\author{Chang Liu$^{\star\dag}$, Wenzhong Yan$^{\star}$, and Ankur Mehta
\thanks{Chang Liu and Ankur Mehta are with the Samueli School of Engineering, Electrical and Computer Engineering, Wenzhong Yan is with the Samueli School of Engineering, Mechanical and Aerospace Engineering, University of California, Los Angeles, Los Angeles, CA, USA
\newline
\indent Programming package can be found at \href{https://git.uclalemur.com/Chang_Liu/Orthogonally_Assembled_Double_Layer_Corrugation/tree/master/2_Modeling/_Optimization_Design_Programming_Package}{\textit{\textcolor{black}{https://git.uclalemur.com/
Chang\_Liu/Orthogonally\_Assembled\_Double\_Layer\_Corrugation/tree/
master/2\_Modeling/\_Optimization\_Design\_Programming\_Package}}}
\newline
\indent $^{\star}$Chang Liu and Wenzhong Yan contributed equally to this work
\newline
\indent $^{\dag}$Corresponding author, {\tt\footnotesize changliu498@ucla.edu}
\newline
\indent This work has been submitted to the IEEE for possible publication. Copyright may be transferred without notice, after which this version may no longer be accessible.
}
}

\begin{document}

\maketitle



\begin{abstract}

Orthogonally assembled double-layered corrugated (OADLC) mechanisms are a class of foldable structures that harness origami-inspired methods to enhance the structural stiffness of resulting devices; these mechanisms have extensive applications due to their lightweight, compact nature as well as their high strength-to-weight ratio. However, the design of these mechanisms remains challenging. Here, we propose an efficient method to rapidly design OADLC mechanisms from desired behavioral specifications, i.e. in-plane stiffness and out-of-plane stiffness. Based on an equivalent plate model, we develop and validate analytical formulas for the behavioral specifications of OADLC mechanisms; the analytical formulas can be described as expressions of design parameters. On the basis of the analytical expressions, we formulate the design of OADLC mechanisms from behavioral specifications into an optimization problem that minimizes the weight with given design constraints. The 2D folding patterns of the optimized OADLC mechanisms can be generated automatically and directly delivered for fabrication. Our rapid design method is demonstrated by developing stiffness-enhanced mechanisms with a desired out-of-plane stiffness for a foldable gripper that enables a blimp to perch steadily under air disturbance and weight limit.

\end{abstract}


\section{Introduction}

Inspired by nature, origami-inspired folding as a top-down design and fabrication strategy has been introduced to create three dimensional (3D) devices and robots from 2D planar materials \cite{onal2014origami}. These folded devices have many advantages, such as tunable mechanical properties \cite{zhai2018origami}, ease of design \cite{mehta2014end,mehta2016design}, inexpensive and rapid manufacturing \cite{onal2015origami,yan2018towards, yan2019_iros}, high scalability \cite{tachi2010origamizing, rogers2016origami, gilewski2014origami}, and low weight \cite{schenk2011origami,liu2017self, li2017fluid, liu2019self}. Therefore, using folding-inspired techniques as a design and fabrication paradigm for developing 3D structures is promising.

\begin{figure}[t]
    \centering
    \includegraphics[width=3.25in]{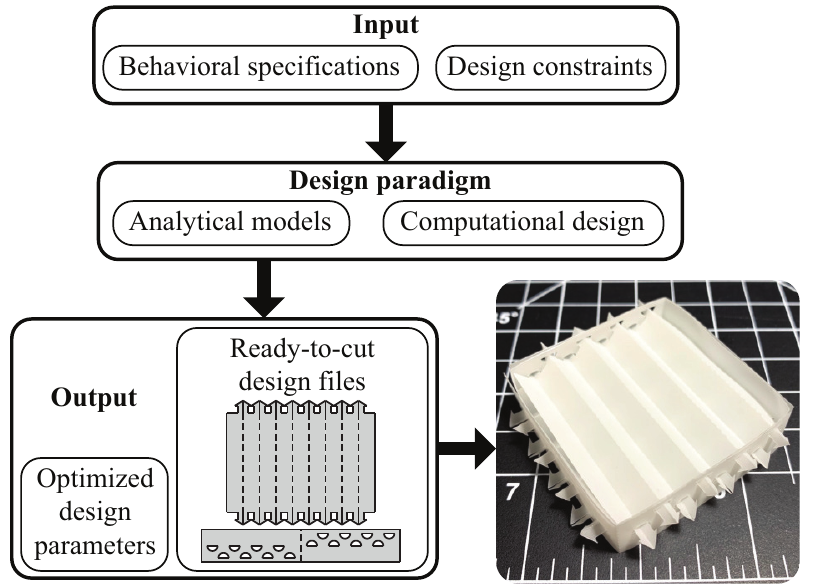}
    \caption{The flow chart of our computational design and fabrication of OADLC mechanisms. The width of the square grid is \SI{12.7}{\milli\meter}.}
    \label{fig: Intro_Figure}
\end{figure}

However, lack of structural stiffness has limited the practical usages of origami-inspired foldable designs \cite{felton2014method, sun2015self}. Therefore, ensuring sufficient stiffness while maintaining  origami-inspired structures' inherent advantages of low weight, ease of design, and inexpensive, rapid fabrication remains a challenge.

Corrugated laminates is one of the most promising solutions for addressing the lack of stiffness while maintaining low weight. Corrugated laminates have extreme anisotropic behavior resulting from folding \cite{yokozeki2006mechanical}, where they have increased stiffness along spanwise direction (along creases) but relative compliance in the chordwise direction (along corrugations), with respect to the constitutive material. 
One way to achieve high stiffnesses along both spanwise and chordwise directions is to assemble two corrugated layers orthogonally into an orthogonally assembled double-layered corrugated (OADLC) mechanism \cite{cheon2015equivalent}. 

There are various models developed to describe the behaviors of single-layered corrugated structures.
Among many homogenization models \cite{briassoulis1986equivalent, andrianov2009asymptotic, arkhangel2007effective, ye2014equivalent, xia2012equivalent}, Xia \textit{et al.} \cite{xia2012equivalent} presented mathematical expressions of stiffness matrices for arbitrary corrugated geometry with easy formulations and high accuracy. 
Models for structures with multi-layered corrugations have also been studied. For manufacturing simplicity, multi-layered corrugations are usually parallelly assembled (creases in all corrugated layers are parallel to each other) \cite{greenfield2017system}.
For example, an analytic homogenization model for parallelly assembled double corrugated core cardboards is presented in \cite{duong2017analysis}. 
New techniques have been developed to manufacture orthogonally assembled corrugated structures \cite{greenfield2017system, greenfield2019system} and different models have been applied. In \cite{cheon2015equivalent}, corrugated layers within corrugated-core sandwich panels were treated and modeled as equivalent continuum layers. However, this method is not intuitive and is limited to structures with relative thin constitutive material compared to the thickness of the corrugation. 
Plenty of research have been done on characterizing behaviors of corrugated laminates. However, the design and construction of such OADLC mechanisms with desired behavioral specifications, e.g. structural stiffness, is still bottlenecked by the design process, which involves numerous iterations of computationally expensive analysis. In order to efficiently design and customize this class of stiffness-enhanced structures in a rapid prototyping manner, a systematic design method needs to be developed.

Here, we present an efficient formulation of OADLC mechanisms from behavioral specifications as a constrained optimization problem, as shown in Fig. \ref{fig: Intro_Figure}. 
Our method is based on an equivalent plate model that converts the corrugated structure into a plate with equivalent elastic constants (Section \ref{section: Modeling and Validation}). 
Thus, we develop analytical expressions of the in-plane stiffness and out-of-plane stiffness of this mechanism (Section \ref{section: Modeling and Validation}). 
Based on these analytical expressions, we can transform the design of OADLC mechanisms with specified structural stiffnesses into a set of constraints on the design parameters. To determine a specific parameter assignment, we can apply these  constraints to an optimization criterion.
For example, incrementing weight usually increases cost and power consumption, reduces mobility \cite{junyao2009study, liu2017three}, and even leads to functional incompetence or failure \cite{sun2015self}, especially for aerial applications \cite{kim2018origami}.
Thus, we choose to minimize the weight of the resulting designs while satisfying stiffness requirements. Other optimization targets, e.g. robustness of the resulting design to manufacturing tolerances \cite{yan2019_iros}, can also be easily adopted. The optimized design can be output instantly from our programming package as ready-to-cut 2D mechanical drawings that can be directly fabricated by 2D machinery, e.g. paper cutter (Section.~\ref{section:ComputationalDesign}). 
Eventually, this optimization-based approach for the rapid design of OADLC mechanisms is demonstrated through designing stiffness-enhanced mechanisms for a foldable gripper on a blimp for steady perching (Section~\ref{section:caseStudy}).

To summarize, the contributions of this work include:
\begin{itemize}
    \item computationally tractable analytical models that characterize the stiffnesses (both in-plane stiffness and out-of-plane stiffness) of OADLC mechanisms,
    \item experimental validation of the analytical stiffness models of OADLC mechanisms,
    \item an optimization-based method to computationally design OADLC mechanisms with desired behavioral specifications while minimizing the weight of designs,
    \item a programming package that can automatically generate ready-to-cut patterns of the optimized design from the specifications, and 
    \item one example that demonstrates the proposed design paradigm.
\end{itemize}


\section{Modeling and Validation}
\label{section: Modeling and Validation}
 
\begin{figure}[t]
    \centering
    \includegraphics[width=3.25in]{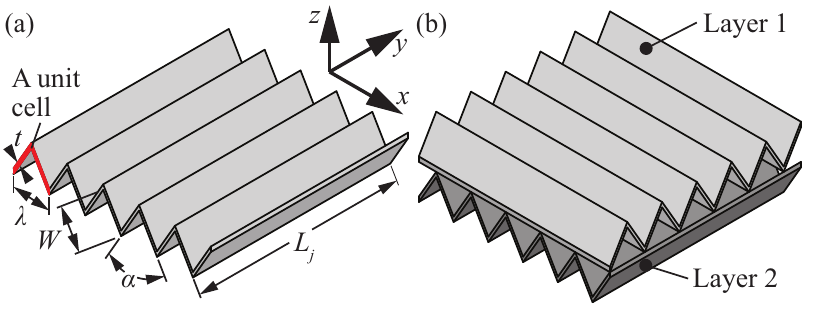}
    \caption{Corrugated mechanisms with triangular unit cell. (a) A single-layered corrugated sheet with it geometry parameters and coordinate system labeled. (b) A typical OADLC mechanism.}
    \label{fig: Corrugation_Design}
\end{figure}

A typical OADLC mechanism is composed of two orthogonally assembled single-layered corrugated structures (Fig \ref{fig: Corrugation_Design}). Thus, we first introduce relevant background knowledge on the modeling of single-layered corrugated structures. Based on this modeling, we then further derive models for the OADLC mechanism. We eventually validate our models with physical experiments.

\subsection{Parameter Definition and Assumptions}
\label{section: Parameter Definitions and Assumptions}

To simplify and analytically derive stiffness models of single-layered corrugated structures, we assume:

\begin{itemize}
    \item the constitutive material is isotropic,
    \item the constitutive material's thickness is constant,
    \item all unit cells (Fig.~\ref{fig: Corrugation_Design}(a)) are identical, and
    \item each single-layered corrugated structure is fabricated symmetrically with respect to its mid-plane.
\end{itemize}


\subsection{Background: Single-Layered Corrugated Structure}
\label{section: Background}

\subsubsection{Generalized Stiffness Matrix}
\label{section: Derivation of Generalized Stiffness Matrix}

With the assumptions mentioned above, the single-layered corrugated structure can be approximated as an orthotropic classical Kirchhoff plate, with the relationship between stress and strain expressed as \cite{huber1923theorie, koudzari2019corrugated}:
\begin{equation}
    \begin{Bmatrix}
        \textbf{N} \\
        \textbf{M}
    \end{Bmatrix}
    =
    \begin{bmatrix}
        \textbf{A} & 0 \\
        0 & \textbf{D}
    \end{bmatrix}
    \begin{Bmatrix}
        \pmb{\epsilon} \\
        \pmb{\kappa}
    \end{Bmatrix}
    \label{Eq: constitutive equation}
\end{equation}

\noindent where $\textbf{N}$, $\textbf{M}$, $\pmb{\epsilon}$, and $\pmb{\kappa}$ are the in-plane force vector, out-of-plane moment vector, membrane strain vector, and curvature strain vector, respectively. $\textbf{A}$ and $\textbf{D}$ are the in-plane stiffness matrix and out-of-plane stiffness matrix, and can be expressed as follows \cite{xia2012equivalent, dayyani2015mechanics}:

\begin{equation}
    \textbf{A} =
    \begin{bmatrix}
        A_{11} & A_{12} & 0 \\
        A_{21} & A_{22} & 0 \\
        0 & 0 & A_{66} \\
    \end{bmatrix}
    ,\hspace{0.3cm}
    \textbf{D} =
    \begin{bmatrix}
        D_{11} & D_{12} & 0 \\
        D_{21} & D_{22} & 0 \\
        0 & 0 & D_{66} \\
    \end{bmatrix}
\end{equation}


\subsubsection{Equivalent Plate Conversion}
\label{section: Derivation of Equivalent Plate Models}

\begin{figure}[t]
    \centering
    \includegraphics[width=3.25in]{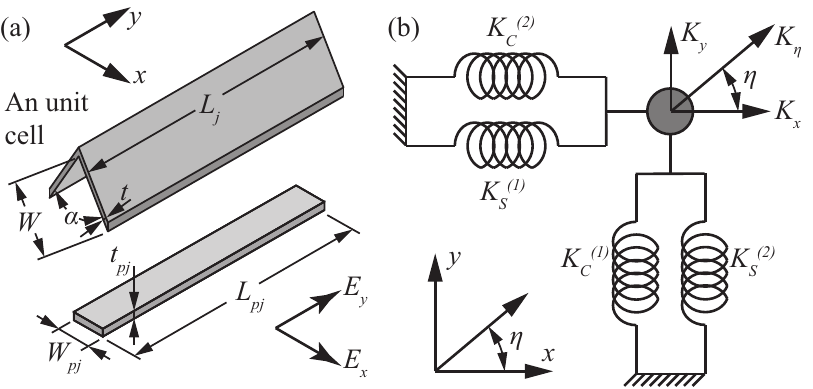}
    \caption{Modeling illustration. (a) Equivalent plate model for a unit cell of a triangular corrugated mechanism. (b) Spring model for an OADLC mechanism. 
    $\eta$ is an arbitrary direction of interest.}
    \label{fig: Equivalent_Plate_Model_Spring_Model}
\end{figure}

Each unit cell (e.g. the $j^{th}$ unit, Fig.~\ref{fig: Equivalent_Plate_Model_Spring_Model}(a)) of the corrugation mechanism can be further simplified as an equivalent plate. This plate's dimension is defined as: thickness $t_{p} = t$, length $L_{pj} = L_j$, and width $W_{p} = W \sin{\frac{\alpha}{2}}$, where $t$ is the thickness of the constitutive material, and $L_j$, $W$, and $\alpha$ are the length, width, and fold angle, respectively, of the $j^{th}$ unit. Thus, the in-plane Young's modulus $E_{cx}$ and $E_{cy}$, and out-of-plane Young's modulus $E_{bx}$ and $E_{by}$ of the equivalent plate are expressed as follows \cite{dayyani2015mechanics}:
\begin{equation}
    E_{cx} = \frac{1}{t} \frac{A_{11} A_{22} - A_{12}^2}{A_{22}}
    ,\hspace{0.3cm}
    E_{cy} = \frac{1}{t} \frac{A_{11} A_{22} - A_{12}^2}{A_{11}}
    \label{eq: E_c}
\end{equation}
\begin{equation}
    E_{bx} = \frac{12 (D_{11} D_{22} - D_{12}^2)}{t^3 D_{22}}
    ,\hspace{0.3cm}
    E_{by} = \frac{12 (D_{11} D_{22} - D_{12}^2)}{t^3 D_{11}}
    \label{eq: E_b}
\end{equation}


Particularly, the equivalent Young's modulus of a triangular unit cell can be derived explicitly by combining its specific expressions of $\textbf{A}$ and $\textbf{D}$\cite{xia2012equivalent} and summarized as follows:
\begin{multline}
    E_{cx} = E \bigg[  t^2 ( \nu^2 - 2\nu - 1) \sin\frac{\alpha}{2} \bigg] \bigg/ \bigg[ (3 \nu^4 + 2 \nu^3 - \nu - 3) t^2 \\ 
    \big(\sin\frac{\alpha}{2} \big)^2 + (-\nu^4 + 2\nu^3 + 4 \nu^2 - 2 \nu - 3) W^2 \cos\frac{\alpha}{2}  \bigg]
    \label{eq: Ecx}
\end{multline} 
\begin{equation}
    E_{cy} = E \frac{ ( \nu^2 - 2 \nu - 3)}{4 (\nu^2 - 1) \sin\frac{\alpha}{2}}
    \label{eq: Ecy}
\end{equation}
\begin{equation}
    E_{bx} = E \frac{W^2 (\cos \alpha + 1) \sin \frac{\alpha}{2} + t^2 (1 - \nu^2) \big(\sin \frac{\alpha}{2}\big)^2 }{2 (1 - \nu^2) \bigg[W^2 \big(\cos \frac{\alpha}{2} \big)^2 + t^2 \big(\sin \frac{\alpha}{2}\big)^2 \bigg] }
    \label{eq: Ebx}
\end{equation}
\begin{equation}
    E_{by} = E \bigg[ \sin\frac{\alpha}{2} + \frac{W^2 \big(\cos \frac{\alpha}{2} \big)^2}{t^2 (1 - \nu^2) \sin\frac{\alpha}{2}} \bigg]
    \label{eq: Eby}
\end{equation}
\noindent where $E$ is the Young's modulus of the constitutive material and $\nu$ is Poisson's ratio of the constitutive material.


\subsection{Analytical Modeling of OADLC Mechanism}
\label{section:OADLC Mechanism with Triangular Unit Cell}

The OADLC mechanism is built by orthogonally assembling two single-layered corrugated structures with connectors along the edges, but no direct connections between the two layers. To obtain analytical expressions to realize rapid and efficient design, we simplify the model by neglecting the stiffness-enhanced effect of the connectors based on the fact that the stiffness of the latter is relatively small compared to that of the corrugated layers. Thus, each layer can be treated as compression/torsional springs along both the spanwise and chordwise directions. Therefore, the in-plane stiffness and out-of-plane stiffness of the OADLC mechanism can be computed using equivalent spring theory.


\subsubsection{Analytical Model of In-Plane Stiffness}
\label{section: in-plane stiffness modeling}


Accordingly, we arrive at the model $K = \frac{E A}{L}$, where $K$, $A$, and $L$ are the axial in-plane stiffness, cross-sectional area, and length. On the $i^{th}$ layer, the total in-plane stiffness along the chordwise direction $K_{C}^{(i)}$ is modeled as $n_i$ springs connected in series, and stiffness along the spanwise direction $K_{S}^{(i)}$ is treated as $n_i$ springs connected in parallel:
\begin{equation}
    K_{C}^{(i)} = \frac{1}{ \sum_{j=1}^{n_i} \frac{W_i \sin{\frac{\alpha_i}{2}}}{E_{cx}^{(i)} L_{i,j} t_i} }
    ,\hspace{0.3cm}
    K_{S}^{(i)} = \sum_{j=1}^{n_i} \frac{E_{cy}^{(i)} W_i t_i \sin{\frac{\alpha_i}{2}}}{L_{i,j}}
    \label{eq: K_C_K_S}
\end{equation}

When two corrugated layers are assembled orthogonally, the in-plane stiffness along an arbitrary direction $\eta$ can be simplified as springs connected as shown in Fig.~\ref{fig: Equivalent_Plate_Model_Spring_Model}(b).
\begin{equation}
    K_{\eta}  = (K_{S}^{(1)} + K_{C}^{(2)}) (\cos{\eta})^2 + (K_{S}^{(2)} + K_{C}^{(1)}) (\sin{\eta})^2
    \label{eq: equivalent_K_eta}
\end{equation}


\subsubsection{Analytical Model of Out-of-Plane Stiffness}
\label{section: Analytical Model of Out-of-Plane Stiffness}


$D = \frac{E I}{L}$, where $D$, $I$, and $L$ are the out-of-plane stiffness, second moment of inertia, and length. The out-of-plane stiffness on the $i^{th}$ layer around the chordwise direction $D_C^{(i)}$, spanwise direction $D_S^{(i)}$ and that of the two layered assembly $D_{\eta}$ are derived as:
\begin{equation}
    D_C^{(i)} = \frac{1}{\sum_{j=1}^{n} \frac{12 W_i \sin{\frac{\alpha_i}{2}}}{E_{bx}^{(i)} L_{i,j} t_i^3}}
    ,\hspace{0.3cm}
    D_S^{(i)} = \sum_{j=1}^{n} \frac{E_{by}^{(i)} t_i^3 W_i \sin{\frac{\alpha_i}{2}} }{12 L_{i,j}}
    \label{eq: D_C_D_S}
\end{equation}
\begin{equation}
    D_{\eta} = (D_{S}^{(1)} + D_{C}^{(2)}) (\cos{\eta})^2 + (D_{S}^{(2)} + D_{C}^{(1)}) (\sin{\eta})^2
    \label{eq: equivalent_D_eta}
\end{equation}

In summary, we can analytically determine both the in-plane stiffness and out-of-plane stiffness with system design parameters as follows: 
\begin{equation}
\begin{split}
    K = & \mathcal{F} (\{E_1, \nu_1, t_1, W_1, n_1, L_{1}, \alpha_1\}, \\
        & \{E_2, \nu_2, t_2, W_2, n_2, L_{2}, \alpha_2\}, \eta) 
    \label{eq: K_sum}
\end{split}
\end{equation}
\begin{equation}
\begin{split}
    D = & \mathcal{G} (\{E_1, \nu_1, t_1, W_1, n_1, L_{1}, \alpha_1\}, \\
        & \{E_2, \nu_2, t_2, W_2, n_2, L_{2}, \alpha_2\}, \eta) 
    \label{eq: D_sum}
\end{split}
\end{equation}
\noindent where the expressions $\mathcal{F}$ and $\mathcal{G}$ can be obtained by combining Eq.~\ref{eq: Ecx} with Eq.~\ref{eq: equivalent_D_eta}. It is worth noting that $L_1$ and $L_2$ are arrays of crease length. 

The stiffness of the OADLC layers is usually much larger than that of the connectors, which can guarantee the accuracy of our models. However, the stiffness of the connector needs to be considered when it becomes comparable to that of OADLC layers. In this paper, we focus on the modeling of OADLC mechanism with triangular unit cells. Our method is applicable to OADLC mechanisms with other types of unit cells, e.g. sinusoidal profile, by merely replacing Eq. \ref{eq: Ecx} through Eq. \ref{eq: Eby} correspondingly.


\subsection{Model Validation for OADLC Mechanism}
\label{section: Model Validation for OADLC Mechanism with Triangular Unit Cell}


To validate our models, we designed experimental tests for both the in-plane and out-of-plane stiffness of OADLC mechanisms. We used Grafix Dura-Lar film as the constitutive material with a measured thickness of \SI{0.125}{\milli\meter}, a Young's modulus of \SI{2.7}{GPa} and a Poisson's ratio of 0.43.
Also, for simplification, we used square OADLC samples (with uniform crease lengths that can be expressed as a function of $W$, $n$ and $\alpha$) and only measured the stiffness along the axial direction (i.e. $\eta=\SI{0}{\degree}$).
Thus, we varied the remaining design parameters, namely the plate width $W$, number of creases $n$, and crease fold angle $\alpha$ (see Eq. \ref{eq: K_sum} and Eq. \ref{eq: D_sum}). In each test, we fixed two out of the three design parameters, and varied the other one to measure stiffness.
Both in-plane compression tests and three-point out-of-plane bending tests were performed using the BAOSHISHAN digital force gauge HP-500 with an HPB manual test stand.

\begin{figure}[t]
    \centering
    \includegraphics[width=3.25in]{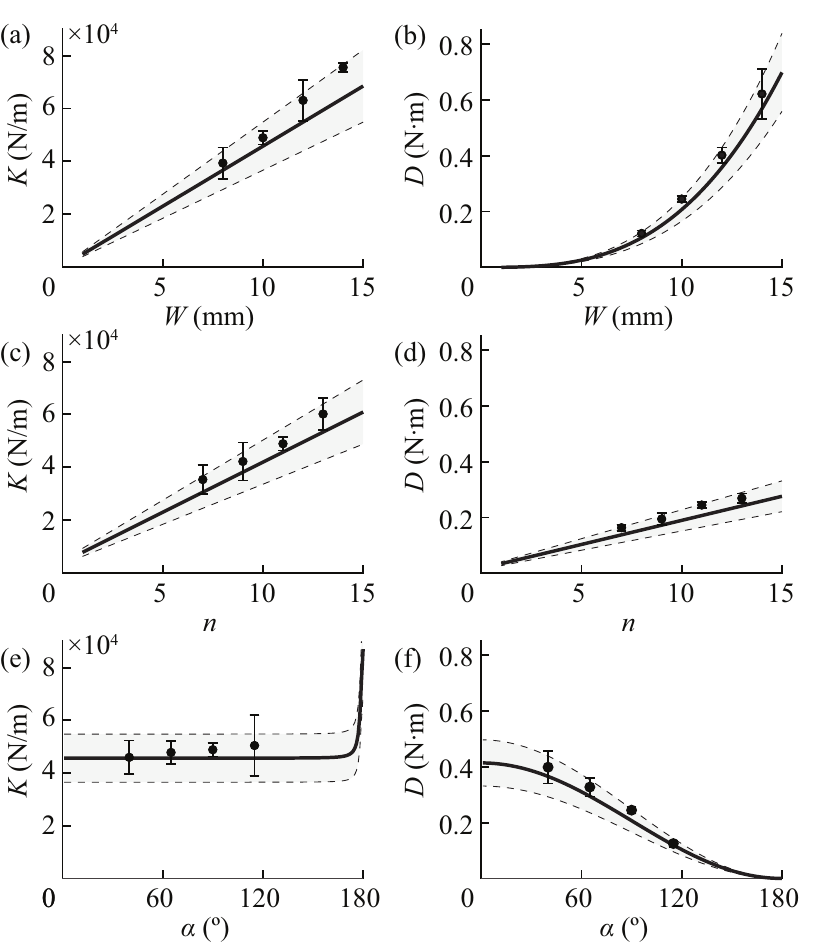}
    \caption{Modeling validation. (a-b) Varying plate width $W$ while fixing number of creases $n$ and crease fold angle $\alpha$. (a) In-plane compression stiffness. (b) Out-of-plane bending stiffness. (c-d) Varying number of creases $n$ while fixing plate width $W$ and crease fold angle $\alpha$. (c) In-plane compression stiffness. (d) Out-of-plane bending stiffness. (e-f) Varying crease fold angle $\alpha$ while fixing plate width $W$ and number of creases $n$. (e) In-plane compression stiffness. (f) Out-of-plane bending stiffness. Dots and error bars indicate mean testing results and standard deviations ($N=3$), solid lines indicate modeling results, dashed lines indicate 20\% above and below modeling results and shaded regions indicate 20\% error ranges.}
    \label{fig: Model_Validation}
\end{figure}

To validate the effect of plate width $W$, we fixed the value of the number of creases and crease fold angle ($n=11$, $\alpha=\SI{90}{\degree}$), and varied $W$ to be $\SI{8}{\milli\meter}$, $\SI{10}{\milli\meter}$, $\SI{12}{\milli\meter}$, and $\SI{14}{\milli\meter}$. Our results are shown in Fig.~\ref{fig: Model_Validation}(a) and (b), respectively, showing a good agreement between our experiments and models as all errors are less than 20\%.
In a similar manner, we fixed $W$ ($=\SI{10}{\milli\meter}$) and $\alpha$ ($=\SI{90}{\degree}$), and changed $n$ to be 7, 9, 11, and 13. The results of the testing of in-plane compression stiffness and out-of-plane bending stiffness are shown in Fig.~\ref{fig: Model_Validation}(c) and (d), respectively, suggesting our model can accurately predict the behavior of an OADLC mechanism. Similarly, we kept the value of $n$ and $W$ constant ($n=11$, $W=\SI{10}{\milli\meter}$), and altered $\alpha$ to be $\SI{40}{\degree}$, $\SI{65}{\degree}$, $\SI{90}{\degree}$ and $\SI{115}{\degree}$, as shown in Fig.~\ref{fig: Model_Validation}(e) and (f) for in-plane compression and out-of-plane bending, respectively. The results also indicate that our model only generates very small error with regard to the experiment. 

Our model has good agreement with experimental testing among a wide range of parameters, which demonstrates the feasibility of our method. All experimental data was slightly larger than the predictions from our model. This discrepancy is mainly caused by the negligence of connectors during modeling. To further improve the accuracy, we can include the connector into our model at the cost of model complexity.


\section{Computational Design}
\label{section:ComputationalDesign}
Our analytical models of OADLC mechanisms effectively guide the design of the system when a desired stiffness (in-plane/out-of-plane) is specified. Here we present one possible optimization algorithm that finds the set of design parameters that satisfies the predefined constraints and minimizes the weight of the resulting design. This optimization approach allows users to customize their own specific constrained optimization problems, with different parameters, constraints, and/or optimization targets, using the scheme to be discussed. To solve this optimization problem, we also develop a programming package, which can instantly output ready-to-cut design files for fabrication and assembly.


\subsection{Design Parameter}
\label{section: System Inputs}

The design parameters of an OADLC mechanism include the mechanical properties of the constitutive material and the geometries of resulting devices.


\subsubsection{Mechanical Properties}

Young's modulus $E_i$, Poisson's ratio $\nu_i$ and thickness $t_i$. $i$ indicates the layer number ($i$ = 1 or 2). 



\subsubsection{Geometry Parameters}

the number of creases $n_{i}$, crease fold angle $\alpha_i$, and plate width $W_i$. 

For demonstration purposes, we use a square OADLC mechanism, which makes $n_1$ equal to $n_2$. For further simplicity, we use the same constitutive material and identical unit cells for both layers. Therefore, $E_1 = E_2 = E$, $\nu_1 = \nu_2 = \nu$, $t_1 = t_2 = t$, $n_1 = n_2 = n$, $\alpha_1 = \alpha_2 = \alpha$, and $W_1 = W_2 = W$.


\subsection{Design Constraint}
\label{section: design constraint}

There are three types of design constraints: layout constraints, fabrication constraints, and behavioral constraints.
Layout constraints stem from the layout (e.g. square, see Fig. \ref{fig: Intro_Figure}) of the OADLC mechanism. This type of constraint imposes geometric relationships for the lengths of creases. Fabrication constraints come from practical considerations for fabrication and assembly. For example, we may need to confine the overall dimensions of the resulting device to fit in a very limited space. Behavioral constraints originate from the demand on the desired performance of resulting devices. 
For instance, users can require that the deformations of the desired OADLC mechanism under certain loading conditions need to be smaller than certain values. Thus, users can set behavioral constraints to define lower bounds of the stiffness of the resulting structure. Constraints of all types on each design parameter can be specified by users.


\subsubsection{Layout Constraints}
Theoretically, the layout of the OADLC mechanism can be any shape.
For a square OADLC mechanism, the total length of each layer along the chordwise direction when folded should be the same as the crease length. This constraint can be expressed as follows:

\begin{equation}
    L_j = (n + 1) W \sin{\frac{\alpha}{2}}
    \label{eq: rect_geo_constrain_1}
\end{equation}


Other layouts, e.g. circular, triangle, trapezoid or any other arbitrary shapes, can also be defined in the same manner.


\subsubsection{Fabrication Constraints}
\label{section: Fabrication Constraint}

Here we propose three common fabrication constraints from our tests as examples, but users are free to define any other constraints for their needs. 

(i) Maximum fabrication dimension $L_{fab}$: the maximum dimension of each layer when laying flat should be smaller than the maximum fabrication dimension $L_{fab}$. Therefore, $\max{(L_j)} \leq L_{fab}$ and $(n+1) W \leq L_{fab}$.

(ii) Folded axial dimension [$L_{min}$, $L_{max}$]: the resulting OADLC mechanism usually needs to fit into a limited space with lower bound $L_{min}$ and upper bound $L_{max}$. Thus, $L_{min} \leq (n+1) W \sin{\frac{\alpha}{2}} \leq L_{max}$.

(iii) Folded thickness [$t_{min}$, $t_{max}$]: the thickness of the OADLC mechanism should also be constrained within an appropriate range. Therefore, $t_{min} \leq 2 W \cos{\frac{\alpha}{2}} \leq t_{max}$.


\subsubsection{Behavioral Constraints}
\label{section: Behavioral Constraint}
In practice, large deformation can cause serious consequences. Therefore, the OADLC mechanisms usually need to meet a minimum stiffness requirement ($K_{min}$ and/or $D_{min}$). Therefore, we have $K_{OADLC} \geq K_{min}$ and/or $D_{OADLC} \geq D_{min}$.


\subsection{Optimization}
\label{section:optimization}

Here, we only use out-of-plane stiffness with a square layout as an example to demonstrate how we formulate our design problem into a well-defined constrained optimization problem. Also, we choose the design with the least weight. 
This formulation also applies to designs using in-plane stiffness or other more sophisticated situations thanks to the available analytical models. Along with the design constraints as discussed in Section \ref{section: design constraint}, the constrained optimization problem can be expressed as follows:

\begin{equation}
   \begin{aligned}
   & \underset{E, \nu, t, \alpha, L, W, n}{\text{minimize}}
   && m = \mathcal{T} (t, \rho, \alpha, W, n)\\
   & \text{subject to}
   && L_j = (n + 1) W \sin{\frac{\alpha}{2}} \\
   &&& (n+1) W \leq L_{fab} \\
   &&& L_{min} \leq (n + 1) W \sin{\frac{\alpha}{2}} \leq L_{max} \\
   &&& t_{min} \leq 2 W \cos{\frac{\alpha}{2}} \leq t_{max} \\
   &&& D_{OADLC} \geq D_{min}
   \end{aligned}
   \label{eq:optimi}
\end{equation}

\noindent where $\rho$ is the constitutive material's density. $m$ is the total weight of the OADLC mechanism, which can be expressed as a function $\mathcal{T}$ of its geometry and material properties.


\section{Case Study}
\label{section:caseStudy}


\begin{figure}[t]
    \centering
    \includegraphics[width=3.25in]{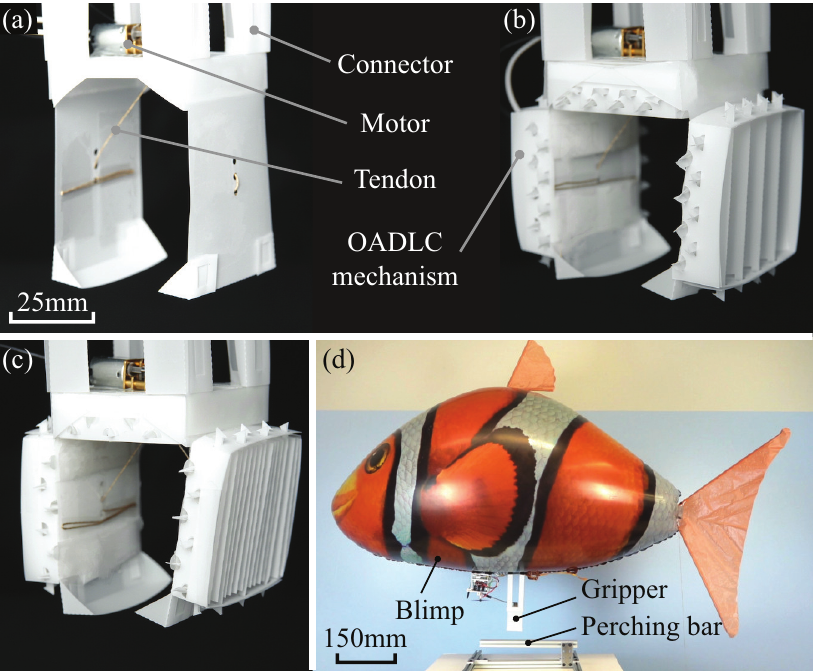}
    \caption{Origami grippers on a blimp. (a) Original gripper, \SI{25.2}{\gram}. (b) Gripper with optimized stiffness enhancement, \SI{30.7}{\gram}. (c) Gripper with naive stiffness enhancement, \SI{37.7}{\gram}. (d) A blimp with the original gripper attached.}
    \label{fig: Blimp_Gripper_Intro}
\end{figure}

Given the limited onboard power supply, the capability of perching is essential for small aerial robots, allowing them to land on trees, walls, or charger lines to rest or recharge. One of the major challenges for perching is to design a light-weight and powerful mechanism to guarantee steady perching \cite{zhang_compliant_2019}. An origami gripper is a promising solution due to its intrinsic low weight. Here, we use an origami gripper (Fig.~\ref{fig: Blimp_Gripper_Intro}(a)) as an example to demonstrate our method to improve its rigidity with desired out-of-plane stiffness while minimizing its weight for the stable perching of a blimp in moderate wind ($\approx$ \SI{40}{cfm} flow rate, \SI{0.25}{\meter} away from the blimp).

The origami gripper (Fig.~\ref{fig: Blimp_Gripper_Intro}(a), \SI{25.2}{\gram}, including a motor and a connector) has two flat fingers attached on the base, which are tendon-driven by a micro gearmotor (Pololu \#2366). 
Initially, the blimp with the original gripper is able to freely fly to reach the perching bar and rest in standing air (Fig.~\ref{fig: Blimp_Demo}(a)). However, due to the lack of bending rigidity, the gripper is unable to stay on the perching bar in a moderate wind environment ($\approx$ \SI{40}{cfm} flow rate, \SI{0.5}{\meter} from the blimp). To improve the perching performance, we propose to increase the bending stiffness of the gripper by adding square OADLC mechanisms onto the fingers and base. Also, due to the limited payload (\SI{31}{\gram}) of the blimp, we need to minimize the weight of the resulting gripper.

\subsection{Constraints}

\begin{itemize}
    \item $L_{fab}=\SI{250}{\milli\meter}$, the maximum working space of the fabrication tool (Silhouette Cameo 4),
    \item $L_{min} = \SI{48}{\milli\meter}$ and $L_{max} = \SI{60}{\milli\meter}$ for matching finger dimensions and ease of mounting,
    \item $t_{min} = \SI{10}{\milli\meter}$ and $t_{max} = \SI{12.5}{\milli\meter}$ for fitting in space under the base without influencing gripper folding, and
    \item $D_{min} = \SI{80}{\milli\newton\cdot\meter}$, guarantee that the deflection of the gripper tip is less than \SI{0.5}{\milli\meter} for steady perching.
\end{itemize}


\subsection{Computational Design}
\label{section: rectangular design}
By inputting design constraints to our programming package, the optimized design with the minimum weight is calculated with its 2D folding patterns (Fig.~\ref{fig: Intro_Figure}); corresponding design parameters are listed in Table~\ref{table: rectangle min weight validation}, Case 1. Also, the square OADLC structure is $\SI{48.18}{\milli\meter} \times \SI{48.18}{\milli\meter}$ with $t_d = \SI{11.89}{\milli\meter}$. All design constraints are satisfied. 

Particularly, this set of optimal design parameters is obtained nearly instantly by executing the optimization package on a typical personal computer, while conventional parameter exploration may have been extremely time-consuming and labor-intensive.



\begin{table}[t]
\centering
\caption{Example test cases that demonstrate design feasibility.}
\centering
\begin{tabular}{ccccccc}
\noalign{\smallskip} \hline \hline \noalign{\smallskip}
& \multirow{2}{*}{Case} & $W$ & \multirow{2}{*}{$n$} & $\alpha$ & $D$ & $m$ \\
& & (mm) &  & ($\SI{}{\degree}$) & ($\SI{}{\milli\newton\cdot\meter}$) & (g) \\
\noalign{\smallskip} \hline \noalign{\smallskip}
Optimized & \textbf{1} & \textbf{8} & \textbf{8} & \textbf{84} & \textbf{87.80} & \textbf{1.85} \\
\noalign{\smallskip} \hline \noalign{\smallskip}
\multirow{5}{*}{Naive} & \textbf{2} & \textbf{6} & \textbf{31} & \textbf{29} & \textbf{223.56}                        & \textbf{4.17} \\
                       & 3 & 6 & 37 & 25 & 269.96 & 4.91 \\
                       & 4 & 6 & 35 & 32 & 247.94 & 5.65 \\
                       & 5 & 6 & 40 & 30 & 287.71 & 6.28 \\
\noalign{\smallskip} \hline \hline \noalign{\smallskip}
\end{tabular}
\label{table: rectangle min weight validation}
\end{table}






\subsection{Validation}


\begin{figure}[b]
    \centering
    \includegraphics[width=3.25in]{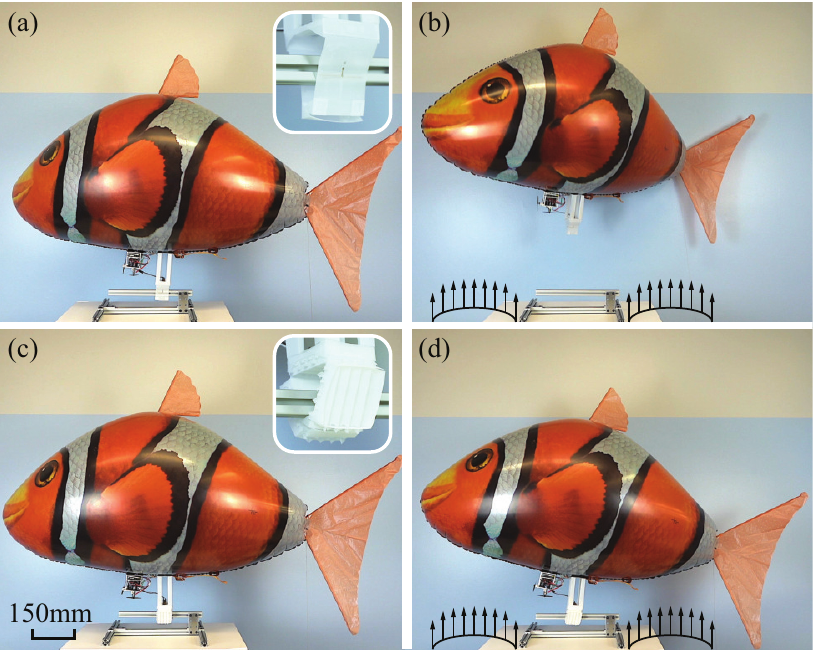}
    \caption{Perching of a blimp using an origami gripper. (a) Blimp can stay on the bar using the original gripper when there is no air disturbance. (b) Blimp detaches the perching bar when air flow increases to \SI{40}{cfm}. (c) Blimp rests on the bar. (d) Blimp stays on the bar even with \SI{40}{cfm} air blowing.}
    \label{fig: Blimp_Demo}
\end{figure}

After attaching the optimized OADLC mechanisms onto the fingers and base, an upgraded gripper was obtained as shown in Fig.~\ref{fig: Blimp_Gripper_Intro}(b) (\SI{30.7}{\gram}, within the limit of blimp's payload). The blimp could still move freely to reach and grasp the perching bar (Fig.~\ref{fig: Blimp_Demo}(c)) and successfully stayed on the bar under the disturbance of the same moderate wind (Fig.~\ref{fig: Blimp_Demo}(d)).

Without using our design tool, it takes an enormous amount of iterations and time to design an OADLC mechanism that satisfies all the design constraints. For comparison, we randomly select several OADLC mechanism designs that had already satisfied all the design constraints through a trial-and-error process without their weight minimized (Table~\ref{table: rectangle min weight validation}, Case 2-5). We chose to fabricate the naive design with the lightest weight, Case 2, resulting in a $\SI{37.7}{\gram}$ gripper. This over-weight gripper caused the blimp fail to lift-off although it has sufficient stiffness for perching under a disturbance.


\subsection{Other Applications}

\begin{figure}[t]
    \centering
    \includegraphics[width=3.25in]{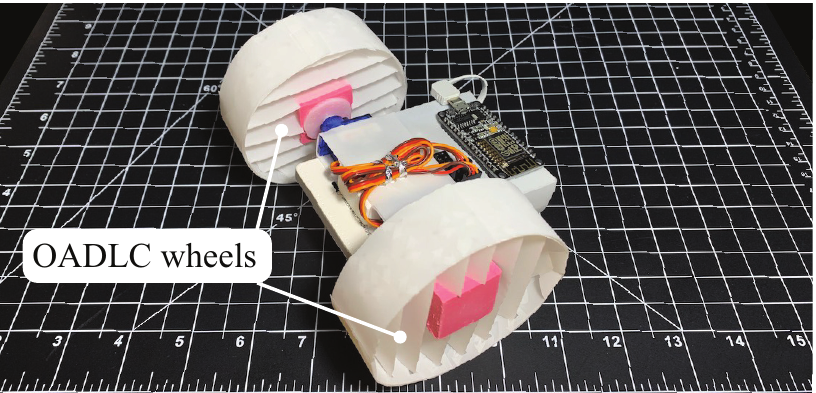}
    \caption{A foldable car with OADLC wheels, proving the design paradigm can be applied to build circular OADLC mechanisms with in-plane stiffness specification as a design constraint. The width of the square grid is \SI{12.7}{\milli\meter}.}
    \label{fig: foldable_car}
\end{figure}

This computational design paradigm can also be applied to design other shapes of OADLC mechanisms with different mechanical specifications. For example, an origami-inspired foldable car derived from author's previous work \cite{mehta2014cogeneration} has been enhanced to gain more load-bearing capacity by replacing laminated wheels with computationally designed circular OADLC wheels (Fig.~\ref{fig: foldable_car}). We believe this design paradigm can be extended to more applications in robotic systems.

\section{Conclusion}
\label{section:conclusion}


We have proposed a computational design and fabrication method for orthogonally assembled double-layered corrugated mechanisms, an origami-inspired stiffness-enhanced design.
The value of this design paradigm
is demonstrated by creating optimized stiffness-enhanced mechanisms using the design tool presented above and adding them onto a robotic system. This system fails to function properly without the mechanisms and also fails when using un-optimized solutions. 
In a matter of seconds, the high-level behavioral specifications of a robotic device can be realized instantly and automatically into design cut files ready to use for fabrication.
This system brings the goal of personal, customized, and on-demand design of stiffness-enhanced mechanisms within reach.

This design framework is one step towards the high-level goal of the computational design of fully functional robotic systems based on user-specified behaviors. 
This work suggests an important next step towards computational foldable robot design: developing design tools to identify the weak points in foldable robotic systems, thus knowing where to apply stiffness-enhanced mechanisms to improve system performance.

Beyond the scope of the proposed corrugation mechanism, our design approach can be applied to other foldable structures thanks to the intrinsic simplicity of origami structures. For example, we can computationally design actuation mechanisms from desired power output specifications by using a similar method to our framework. Furthermore, our method holds the potential to be adopted to achieve rapid design and prototyping of dynamic reconfigurable systems and integrated origami robots from behavioral specifications.


\section*{acknowledgment}

This work is partially supported by the National Science Foundation (grants \#1752575).
The authors would like to thank Jiahao Li, Ryan Chen, and Ethan Uetrecht for their assistance.

\addtolength{\textheight}{-5.9cm}


\bibliographystyle{IEEEtran}
\bibliography{Reference}


\end{document}